\documentclass{article} 
\usepackage{iclr2023_conference,times}

\usepackage{amsmath,amsfonts,bm}

\def\eqref#1{equation~\ref{#1}}

\def\1{\bm{1}}

\DeclareMathAlphabet{\mathsfit}{\encodingdefault}{\sfdefault}{m}{sl}
\SetMathAlphabet{\mathsfit}{bold}{\encodingdefault}{\sfdefault}{bx}{n}

\usepackage{hyperref}
\usepackage{url}
\usepackage{float}

\newcommand{\proposed}{\textsf{DOSTransformer}}

\title{Predicting Density of States via Multi-modal Transformer}
\iclrfinalcopy

\author{Namkyeong Lee$^{1{\dagger}}$, Heewoong Noh$^{1{\dagger}}$, Sungwon Kim$^{1}$, Dongmin Hyun$^{2}$, Gyoung S. Na$^{3}$, \\
\textbf{Chanyoung Park$^{1}$}\thanks{Corresponding author.~~~{$^\dagger$} These authors contributed equally. 
} \\
$^{1}$ KAIST \hspace{0.15in} 
$^{2}$ POSTECH \hspace{0.15in}
$^{3}$ KRICT \hspace{0.15in} \\
\texttt{\{namkyeong96,heewoongnoh,swkim,cy.park\}@kaist.ac.kr} \\
\texttt{dm.hyun@postech.ac.kr}, \texttt{ngs0@krict.re.kr} \\
}

\usepackage{multirow}
\usepackage{xcolor}
\usepackage{pifont}
\usepackage{graphicx}
\newcommand{\cmark}{\textcolor{blue}{\ding{51}}}%
\newcommand{\xmark}{\textcolor{red}{\ding{55}}}%

\begin{document}

\maketitle

\vspace{-2ex}
\begin{abstract}
\vspace{-2ex}
The density of states (DOS) is a spectral property of materials, which provides fundamental insights on various characteristics of materials.
In this paper, we propose a model to predict the DOS by reflecting the nature of DOS: \textit{DOS determines the general distribution of states as a function of energy.}
Specifically, we integrate the heterogeneous information obtained from the crystal structure and the energies via multi-modal transformer,
thereby modeling the complex relationships between the atoms in the crystal structure, and various energy levels.
Extensive experiments on two types of DOS, i.e., Phonon DOS and Electron DOS, with various real-world scenarios demonstrate the superiority of \proposed. The source code for~\proposed~is available at \url{https://github.com/HeewoongNoh/DOSTransformer}.
\end{abstract}

\vspace{-2ex}
\section{Introduction}
\begin{figure*}[t]
    \centering
  \includegraphics[width=0.94\linewidth]{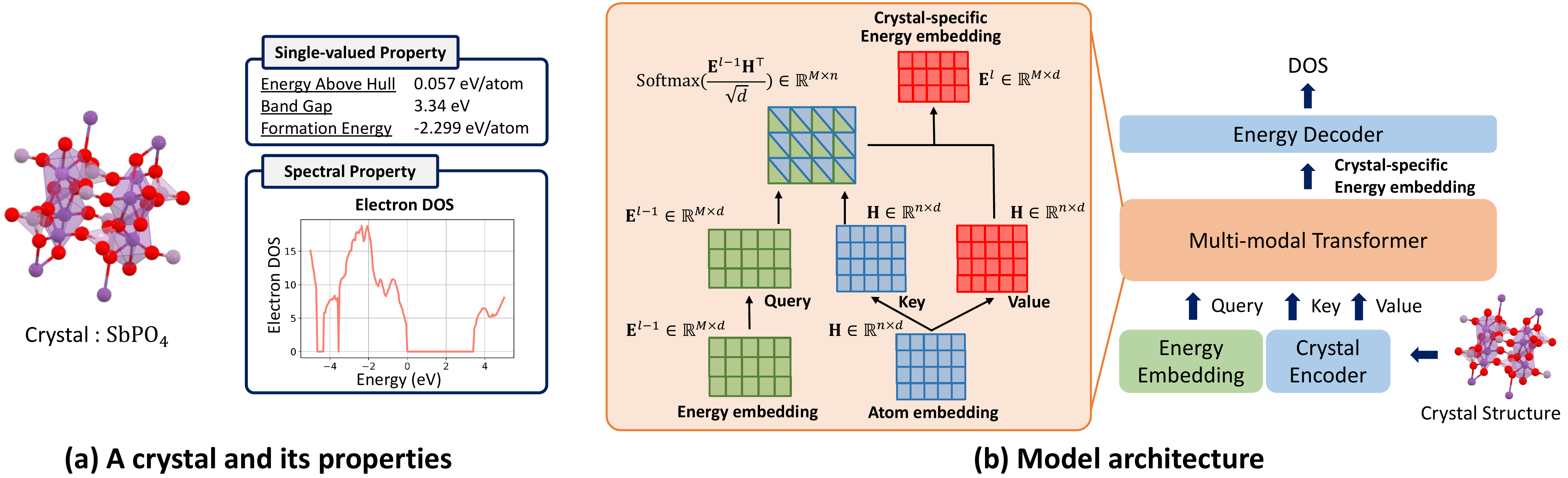}
  \vspace{-1ex}
  \caption{(a) Crystal structure and its various types of properties. (b) Overall model architecture.}
  \label{fig:fig1}
  \vspace{-2ex}
\end{figure*}

\vspace{-2ex}
Despite the recent progress of machine learning (ML) in materials science, most ML models developed in the field have been focused on material properties consisting of single-valued properties \cite{kong2022density}, e.g., band gap energy \cite{lee2016prediction}, formation energy \cite{ward2016general}, and Fermi energy \cite{xie2018crystal}. On the other hand, spectral properties are ubiquitous in materials science, characterizing various properties of materials, e.g., X-ray absorption, dielectric function, and electronic density of states \cite{kong2022density} (See Figure \ref{fig:fig1}(a)).

{
The density of states (DOS), which is the main focus of this paper, is a spectral property that provides fundamental insights on various characteristics of materials, even enabling direct computation of single-valued properties \cite{fung2022physically}.
}
For example, DOS is utilized as a feature of materials for analyzing the underlying reasons for changes in electrical conductivity \cite{deringer2021origins}.
Moreover, band gaps and edge positions, which can be directly derived from DOS, are utilized to discover new photoanodes for solar fuel generation \cite{singh2019robust,yan2017solar}.
Consequently, investigating the ML capability for DOS prediction moves it one step closer to the fundamentals of materials science, thereby accelerating the materials discovery process
However, credible computation of DOS requires expensive time/financial costs of exhaustively conducting experiments with expertise knowledge \cite{chandrasekaran2019solving,del2020efficient}.
Therefore, alternative algorithmic approaches for DOS calculation are necessary, whereas ML capabilities for learning such spectral properties of the crystal structure are relatively under-explored.

Existing studies for DOS prediction with ML models mainly focus on obtaining high-quality representations of crystal structures.
Specifically, \cite{chandrasekaran2019solving,del2020efficient} predict DOS with multi-layered perceptrons (MLPs) given rule-based fingerprints of each grid point and atom, respectively.
Inspired by the recent success of graph neural networks (GNNs) on a variety of tasks in biochemistry \cite{MPNN,stokes2020deep,jiang2021could}, \cite{fung2022physically} leverages GNNs to encode crystal structures to predict DOS with additional physical properties.
Moreover, \cite{chen2021direct} predicts phonon-DOS with euclidean neural networks \cite{thomas2018tensor,kondor2018clebsch,weiler20183d}, which by construction are equivariant to 3D rotations, translations, and inversion, aiming to capture a full crystal symmetry.

Despite their success, existing ML methods for DOS prediction overlook the nature of DOS calculation: \textit{DOS determines the general distribution of states as a function of energy}.
That is, DOS of the crystal structure is determined by not only the structure itself but also the \textit{energy levels}.
Therefore, integrating heterogeneous signals from both the crystal structure and the energy is crucial for DOS prediction, which however has been overlooked by existing studies \cite{chandrasekaran2019solving,del2020efficient,fung2022physically,chen2021direct}.

In this paper, we formulate the DOS prediction problem as a multimodal learning problem, which recently got a surge of interest from ML researchers in various domains thanks to its capability of extracting and relating information from heterogeneous data types \cite{lin2015learning,wang2016learning,bayoudh2020hybrid,baltruvsaitis2018multimodal}.
Specifically, we propose a multimodal transformer model for DOS prediction, named \proposed, which incorporates the crystal structure and the energy as heterogeneous modalities.
{Distinguished from exising studies, \proposed~learns embeddings of energy that are used for modeling complex relationships between the atoms in crystal structure and various energy levels through a cross-attention mechanism.}
By doing so, \proposed~obtains multiple representations for a single crystal structure according to various energy levels, enabling the prediction of a single DOS value on each energy level.

Our extensive experiments on two types of DOS, i.e., Phonon DOS and Electron DOS, and three data split strategies for real-world materials discovery, i.e., one in-distribution split (random split), and two out-of-distribution splits (split according to the number of atom species, and the crystal systems), 
demonstrate the superiority of \proposed~compared with previous methods.
To the best of our knowledge, this is the first work to model the complex relationship between the crystal structure and various energy levels for predicting DOS of the crystal structure.

\section{Preliminaries}
\vspace{-2ex}
\noindent \textbf{Notations.}
Let $\mathcal{G} = (\mathcal{V}, \mathcal{A})$ denote a crystal structure, where $\mathcal{V} = \{v_{1}, \ldots, v_{n}\}$ represents the set of atoms, and $\mathcal{A} \subseteq \mathcal{V} \times \mathcal{V}$ represents the set of edges connecting the atoms in the crystal structure.
Moreover, $\mathcal{G}$ is associated with a feature matrix $\mathbf{X} \in \mathbb{R}^{n \times F}$ and an adjacency matrix $\mathbf{A} \in \mathbb{R}^{n \times n}$ where $\mathbf{A}_{ij} = 1$ if and only if $(v_i, v_j) \in \mathcal{A}$ and $\mathbf{A}_{ij} = 0$ otherwise.

\noindent \textbf{Task: Density of States Prediction.}
Given a set of crystals $\mathcal{D}_{\mathcal{G}} = \{\mathcal{G}_{1}, \mathcal{G}_{2}, \ldots, \mathcal{G}_{N}\}$ and a set of energies $\mathcal{D}_{\mathcal{E}} = \{\mathcal{E}_{1}, \mathcal{E}_{2}, \ldots, \mathcal{E}_{M}\}$, our goal is to train a model $\mathcal{M}$ that predicts the DOS of a crystal structure given a set of energies, i.e., $\mathbf{Y}^{i} = \mathcal{M}(\mathcal{G}_{i}, \mathcal{D}_{\mathcal{E}})$, where $\mathbf{Y}^{i}\in\mathbb{R}^{M}$ is an $M$ dimensional vector containing the DOS values of a crystal structure $\mathcal{G}_{i}$ at each energy $\mathcal{E}_{1}, \ldots, \mathcal{E}_{M}$, and $\mathbf{Y}^{i}_j\in\mathbb{R}$ is the DOS value of $\mathcal{G}_{i}$ at energy level $\mathcal{E}_j$.

\section{Methodology}
\vspace{-2ex}
In this section, we introduce our proposed method named \proposed, a novel DOS prediction framework that learns the complex relationship between the atoms in the crystal structure and various energy levels by utilizing a cross-attention mechanism of the multi-modal transformer.
The overall model architecture is depicted in Figure \ref{fig:fig1} (b).

\subsection{Crystal Encoder}
\vspace{-1ex}
Before modeling the pairwise interaction between the atoms and the energies, we first encode the crystal structure with GNNs to learn the representation of each atom, which contains not only the feature information but also the structural information.
Formally, given a crystal structure $\mathcal{G} = (\mathbf{X}, \mathbf{A})$, we generate an atom embedding matrix for the crystal structure as follows:
\begin{equation}
\small
    \mathbf{H} = \text{GNN} (\mathbf{X}, \mathbf{A}),
    \label{eq: GNN}
\end{equation}
\looseness=-1
where $\mathbf{H} \in \mathbb{R}^{n \times d}$ is an atom embedding matrix for $\mathcal{G}$, whose $i$-th row indicates the representation of atom $v_{i}$, and we stack $L'$ layers of GNNs.
Among various GNNs, we adopt graph networks \cite{battaglia2018relational} as our crystal encoder, which is a generalized and extended version of various GNNs.

\subsection{Multi-modal Transformer}
\vspace{-1ex}
After obtaining the atom embedding matrix $\mathbf{H}$, we model the relationship between the atoms and various energy levels via a cross-attention mechanism of the multi-modal transformer.
{Specifically, we expect the multi-modal transformer to generate the energy-specific representation of the crystal by repeatedly reinforcing the energy representation with the crystal structure.}
To do so, we first introduce a learnable embedding matrix $\mathbf{E}^{0} \in \mathbb{R}^{M \times d}$, whose $j$-th row, i.e., $\mathbf{E}^0_{j}$, indicates the embedding of energy $\mathcal{E}_{j} \in \mathcal{D}_{\mathcal{E}}$.
Then, we present a cross-modal attention for fusing the information from the crystal structure into energy as follows:
\begin{equation}
\small
\begin{split}
    \mathbf{E}^{l} &= \text{Cross-Attention} (\mathbf{Q}_{\mathbf{E}^{l-1}}, \mathbf{K}, \mathbf{V}) \in\mathbb{R}^{M\times d}\\
    &= \text{Softmax} (\frac{\mathbf{E}^{l-1}\mathbf{H}^\top}{\sqrt{d}})\mathbf{H},
\end{split}
\end{equation}
\looseness=-1
where $l = 1,\ldots, L$ indicates the index number of the transformer layer.
In contrast to conventional Transformers, which use learnable weight matrices for query $\mathbf{Q}$, key $\mathbf{K}$, and value $\mathbf{V}$, we directly employ the previously obtained energy embedding matrix $\mathbf{E}^{l-1}$ as the query matrix and the atom embedding matrix $\mathbf{H}$ as the key and value matrices.
Based on the above cross-attention mechanism, we obtain the crystal-specific energy embedding $\mathbf{E}^{l}\in\mathbb{R}^{M\times d}$ by aggregating the information regarding the atoms in the crystal structure that was important at the given energy level.
Consequently, the model learns the crystal-specific energy embedding matrix $\mathbf{E}^{L}\in\mathbb{R}^{M\times d}$ that reflects the complex relationship between the atoms in the crystal structure and various energy levels.

\subsection{Energy Decoder}
\vspace{-1ex}
After obtaining the crystal-specific energy embedding matrix $\mathbf{E}^{L,i}$ of a crystal structure $\mathcal{G}_{i}$, the DOS value at each energy level $\mathcal{E}_j$, i,e., $\hat{\mathbf{Y}}^{i}_{j}$, is given as follows:
\begin{equation}
\small
    \hat{\mathbf{Y}}^{i}_{j} = \phi(\mathbf{E}_{j}^{L,i}  + \alpha \cdot \mathbf{g}_{i}),
    \label{eq: prediction}
\end{equation}
where $\phi: \mathbb{R}^{d} \rightarrow \mathbb{R}^{1}$ is a parameterized MLP for predicting DOS from the given crystal-specific energy embedding of crystal structure $\mathcal{G}_i$ at energy level $j$, i.e., $\mathbf{E}_{j}^{L,i}$ and $\mathbf{g}_{i} \in \mathbb{R}^{d}$, which is a sum pooled representation of crystal $\mathcal{G}_{i}$, and $\alpha$ is a learnable parameter.
Note that $\mathbf{E}_{j}^{L,i}$ indicates $j$-th row of energy embedding matrix $\mathbf{E}_j^{L,i}$.
Finally, \proposed~is trained to minimize the root mean squared error loss $\mathcal{L}$ between the predicted target value $\hat{\mathbf{Y}}_{j}^{i}$ and the ground truth target value $\mathbf{Y}_{j}^{i}$, i.e., $\mathcal{L} = \frac{1}{N \cdot M}\sum_{i = 1}^{N}{\sum_{j = 1}^{M}{\sqrt{(\hat{\mathbf{Y}}^{i}_{j} - \mathbf{Y}^{i}_{j})^{2}}}}$.

\section{Experiments}
\vspace{-1ex}
\subsection{Experimental Setup}
\vspace{-1ex}
\noindent\textbf{Datasets.}
We use two datasets to comprehensively evaluate the performance of \proposed, i.e., Phonon DOS and Electron DOS.
We provide more details on datasets in Appendix \ref{App: Datasets}.

\noindent\textbf{Evaluation Protocol.} 
For Phonon DOS, we evaluate \proposed~with given data splits in a previous work \cite{chen2021direct}.
For Electron DOS, we evaluate \proposed~in three data splits, i.e., one in-distribution split and two out-of-distribution splits.
For in-distribution, we randomly split the dataset into train/valid/test of 80/10/10\%.
On the other hand, for out-of-distribution, we evaluate the model performance on the crystal structures that 1) contain a different number of atom species with the training set, and 2) belong to different crystal systems that were not included in the training set.
Moreover, we predict the Fermi energy of the crystal structure based on the predicted DOS to evaluate how much physically meaningful DOS is predicted by the proposed method.
We provide further details on data split and evaluation on Fermi energy in Appendix \ref{app: Evaluation Protocol}.

\noindent\textbf{Methods Compared.}
We mainly compare \proposed~to recently proposed state-of-the-art method, i.e., E3NN \cite{chen2021direct}.
We also compare \proposed~to simple baseline methods, i.e., MLP and Graph Network \cite{battaglia2018relational}, which predicts the entire DOS sequence directly from the learned representation of the crystal structure.
{
Moreover, to evaluate the effectiveness of the transformer layer that considers the relationship between the atoms and various energy levels, we integrate energy embeddings into baseline methods for DOS prediction as done in Equation \ref{eq: prediction}.
We provide more details on the implementation and compared methods in Appendix \ref{App: Implementation Details} and \ref{App: Methods Compared}, respectively.
}

\noindent\textbf{Evaluation Metrics.}
The performance of \proposed~is mainly evaluated in terms of RMSE and MAE following previous work \cite{chen2021direct}.

\begin{table}[h]
\centering
\vspace{-1ex}
\caption{Overall model performance.} 
\resizebox{0.99\linewidth}{!}{
\renewcommand{\arraystretch}{1}
\begin{tabular}{c|c|cc|ccc||ccc|ccc}
\hline
& \multirow{3}{*}{\begin{tabular}[c]{@{}c@{}}Energy \end{tabular}} & \multicolumn{5}{c||}{In-Distribution} & \multicolumn{6}{c}{Out-of-Distribution (Electron DOS)} \\ \cline{3-13}
&  & \multicolumn{2}{c|}{Phonon DOS} & \multicolumn{3}{c||}{Electron DOS} & \multicolumn{3}{c|}{Scenario 1: \# Atom species} & \multicolumn{3}{c}{Scenario 2: Crystal System} \\ \cline{3-13}
&  & RMSE          & MAE          & RMSE & MAE & Fermi E. & RMSE & MAE & Fermi E. & RMSE & MAE & Fermi E. \\ \hline \hline
\multirow{2}{*}{MLP} & \multirow{2}{*}{\xmark} &0.1719  &0.1131  & 0.2349 & 0.1829 & 2.1314  & 0.2655 & 0.2043 & 2.3852 & 0.2584 & 0.1984 & 2.4863 \\
&  &\scriptsize{(0.0006)}  &\scriptsize{(0.0001)}  & \scriptsize{(0.0008)} & \scriptsize{(0.0009)} & \scriptsize{(0.0908)} & \scriptsize{(0.0025)} & \scriptsize{(0.0022)} & \scriptsize{(0.0916)} & \scriptsize{(0.0026)} & \scriptsize{(0.0028)} & \scriptsize{(0.3264)} \\
\multirow{2}{*}{Graph Network} & \multirow{2}{*}{\xmark} &0.1650  &0.1067  & 0.1529 & 0.1152 & 1.5693  & 0.2225 & 0.1676 & 1.9237 & 0.2041 & 0.1523 & 1.7790 \\
&  &\scriptsize{(0.0030)}  &\scriptsize{(0.0033)}  & \scriptsize{(0.0011)} & \scriptsize{(0.0008)} & \scriptsize{(0.0321)} & \scriptsize{(0.0005)} & \scriptsize{(0.0010)} & \scriptsize{(0.1501)} & \scriptsize{(0.0013)} & \scriptsize{(0.0009)} & \scriptsize{(0.0832)} \\
\multirow{2}{*}{E3NN} & \multirow{2}{*}{\xmark} &0.1356  &0.0809  & 0.1514 & 0.1108 & 1.5790 & 0.2104 & 0.1524 & 1.7308 & 0.1858 & 0.1349 & 1.8642 \\
&  &\scriptsize{(0.0019)}  &\scriptsize{(0.0014)}  & \scriptsize{(0.0013)} & \scriptsize{(0.0009)} & \scriptsize{(0.0415)} & \scriptsize{(0.0007)} & \scriptsize{(0.0007)} & \scriptsize{(0.0357)} & \scriptsize{(0.0006)} & \scriptsize{(0.0004)} & \scriptsize{(0.0169)} \\ \hline
\multirow{2}{*}{MLP} & \multirow{2}{*}{\cmark} &0.1445  & 0.0965 & 0.1604 & 0.1228 & 1.8488  & 0.2080 & 0.1566 & 1.9343 & 0.1905 & 0.1445 & 2.2635 \\
&  &\scriptsize{(0.0000)}  &\scriptsize{(0.0000)}  & \scriptsize{(0.0011)} & \scriptsize{(0.0009)} & \scriptsize{(0.0636)} & \scriptsize{(0.0006)} & \scriptsize{(0.0004)} & \scriptsize{(0.0440)} & \scriptsize{(0.0010)} & \scriptsize{(0.0008)} & \scriptsize{(0.0401)} \\
\multirow{2}{*}{Graph Network} & \multirow{2}{*}{\cmark} &0.1316  &0.0900  & 0.1344 & 0.1000 & 1.5459  & 0.1958 & 0.1451 & 1.7543 & 0.1759 & 0.1297 & 1.7548 \\
&  &\scriptsize{(0.0016)}  &\scriptsize{(0.0008)}  & \scriptsize{(0.0006)} & \scriptsize{(0.0007)} & \scriptsize{(0.0276)} & \scriptsize{(0.0008)} & \scriptsize{(0.0007)} & \scriptsize{(0.0568)} & \scriptsize{(0.0009)} & \scriptsize{(0.0008)} & \scriptsize{(0.0889)} \\
\multirow{2}{*}{E3NN} & \multirow{2}{*}{\cmark} &\textbf{0.1262}  &\textbf{0.0765} & 0.1498 & 0.1109 & 1.6158 & 0.2072 & 0.1540 & 1.9004 & 0.1842 & 0.1348 & 1.8819 \\
&  &\scriptsize{(0.0005)}  & \scriptsize{(0.0008)} & \scriptsize{(0.0008)} & \scriptsize{(0.0008)} & \scriptsize{(0.0311)} & \scriptsize{(0.0005)} & \scriptsize{(0.0016)} & \scriptsize{(0.1281)} & \scriptsize{(0.0005)} & \scriptsize{(0.0007)} & \scriptsize{(0.0732)} \\ \hline
\multirow{2}{*}{\proposed} & \multirow{2}{*}{\cmark} &0.1283  &0.0786  & \textbf{0.1283} & \textbf{0.0918} & \textbf{1.4387} & \textbf{0.1918} & \textbf{0.1373} & \textbf{1.6159} & \textbf{0.1722} & \textbf{0.1231} & \textbf{1.7267} \\
&  & \scriptsize{(0.0017)}  & \scriptsize{(0.0013)}  & \scriptsize{(0.0005)} & \scriptsize{(0.0006)} & \scriptsize{(0.0221)} & \scriptsize{(0.0006)} & \scriptsize{(0.0005)} & \scriptsize{(0.0608)} & \scriptsize{(0.0013)} & \scriptsize{(0.0012)} & \scriptsize{(0.0672)} \\ \hline
\end{tabular}}
\label{tab: main table}
\end{table}
\vspace{-1ex}

\subsection{Experimental Results}
\vspace{-1ex}
The experimental results on two datasets with various evaluation protocols are given in Table \ref{tab: main table}.
We have the following observations:
\textbf{1)} Comparing the baseline methods that overlook the energy levels (i.e., Energy \xmark) with their counterparts that incorporate the energy and the crystal structure as heterogeneous modalities through the energy embeddings (i.e., Energy \cmark), we find out that using the energy embeddings consistently enhances the model performance.
This indicates that making predictions on each energy-level is crucial for DOS prediction, which also aligns with the domain knowledge of materials science, i.e., DOS determines the general distribution of states as a function of energy.
\textbf{2)} On the other hand, \proposed~outperforms previous methods that do not consider the complex relationships between the atoms in crystal structure and various energy levels.
This implies that naively integrating the energy information cannot fully benefit from the energy information.
We further analyze the model performance in terms of the out-of-distribution scenarios, i.e., different atom species numbers and the crystal systems
in Appendix \ref{app: Model Performance Analysis}.
\textbf{3)} Moreover, regarding the capability of predicting the Fermi energy, we observe that predicting the Fermi energy based on the DOS predicted by \proposed~consistently outperforms that of baseline methods.
This indicates that \proposed~predicts physically meaningful DOS, which can further accelerate the materials discovery process.
\textbf{4)} In the case of Phonon DOS, \proposed~performs on par with E3NN with energy embeddings.
This is because the dataset contains a limited number of crystals (1,522 species) compared with the Electron DOS dataset (38,889 species).
However, considering that Electron DOS is much more complex than Phonon DOS in a variety of ways \cite{kong2022density}, and that we are interested in a general prediction method that can be applied to various types of crystals, we argue that \proposed~is practical in the real-world application.

\section{Conclusion}
\vspace{-1ex}
In this paper, we propose \proposed, which predicts the DOS of a crystal structure by modeling the complex relationships between the atoms in the crystal structure and various energy levels.
Extensive experiments verify that incorporating energy information is crucial in predicting the DOS of a crystal structure, and modeling the complex relationship via cross-attention can further improve the model performance.

\section*{Acknowledgements}
This work was supported by Institute of Information \& communications Technology Planning \& Evaluation (IITP) grant funded by the Korea government(MSIT) (No.2022-0-00077), and the core KRICT project from the Korea Research Institute of Chemical Technology (SI2051-10).

\bibliography{iclr2023_conference}
\bibliographystyle{iclr2023_conference}

\appendix
\section{Appendix}

\subsection{Datasets}
\label{App: Datasets}
In this section, we provide further details on the dataset used during training.

\subsubsection{Phonon DOS}
We use the \textbf{Phonon DOS} dataset following the instructions of the official Github repository\footnote{\url{https://github.com/zhantaochen/phonondos_e3nn}} of a previous work \cite{chen2021direct}.
This dataset contains 1,522 crystals whose phonon DOS is calculated from density functional perturbation theory (DFPT) by a previous work \cite{petretto2018high}.
We use the data splits provided in the Github repository to evaluate model performance in the Phonon DOS dataset.


\subsubsection{Electron DOS}
We also use \textbf{Electron DOS} dataset that contains a further variety of crystal structures during training.
Electron DOS dataset consists of the materials and its electron DOS information that are collected from Materials Project (MP) \footnote{\url{https://materialsproject.org/}}.

\noindent \textbf{Data Preprocessing.}
In MP dataset, we exclude the materials that are tagged to include magnetism because the DOS of magnetism materials is not accurate to be directly used for training machine learning models \cite{illas2004extent}.
{We consider an energy grid of 201 points ranging from $-5$ to $5$ eV with respect to the band edges with $50$ meV intervals and the Fermi energy is all set to $0$ eV on this energy grid.}
Moreover, we normalize the DOS of each material to be in the range between 0 and 1. That is, the maximum and minimum value for each DOS is 1 and 0, respectively, for all materials.
Moreover, we smooth the DOS values with the Savitzky-Golay filter with the window size of 17 and polyorder of 1 using scipy library following a previous work \cite{chen2021direct}.

\noindent \textbf{Data Statistics.}
As described in the main manuscript, we further evaluate the model performance in two out-of-distribution scenarios: \textbf{Scenario 1}: regarding the number of atom species, and \textbf{Scenario 2}: regarding the crystal systems.
We provide detailed statistics of the number of crystals for each scenario in Table \ref{app tab: number of elements} and Table \ref{app tab: crystal system}.

\begin{table}[h]
\centering
\caption{The number of crystals according to the number of atom species (Scenario 1).}
\small
\resizebox{0.99 \linewidth}{!}{%
\renewcommand{\arraystretch}{1.1}
\begin{tabular}{c|cccccccc|c}  
\hline
                              &                                                          & \ Unary & \ Binary & Ternary & Quaternary & Quinary & Senary & Septenary & \multirow{2}{*}{\begin{tabular}{l}Total\end{tabular}} \\ 
                              &                                                          & ($1$) & ($2$) & ($3$) & ($4$) & ($5$)& ($6$)& ($7$)\\ 
                              \hline \hline
\multicolumn{1}{c|}{\# Crystals}                                                      &         
&386                 &9,034         &21,794         &5,612           &1,750 &279 &34 &38,889 \\ \hline
\end{tabular}}
\label{app tab: number of elements}
\end{table}

\begin{table}[h]
\centering
\caption{The number of crystals according to different crystal systems (Scenario 2).}
\small
\resizebox{0.99 \linewidth}{!}{%
\renewcommand{\arraystretch}{1.1}
\begin{tabular}{c|cccccccc|c}  
\hline
                              &                                                          &  Cubic &  Hexagonal & Tetragonal & Trigonal & Orthorhombic & Monoclinic & Triclinic &Total \\      
                              \hline \hline
\multicolumn{1}{c|}{\# Crystals}                                                      &         
&8,385                 &3,983         &5,772         &3,964           &8,108 &6,576 &2,101 &38,889 \\ \hline

\end{tabular}}
\label{app tab: crystal system}
\end{table}

\subsection{Evaluation Protocol}
\label{app: Evaluation Protocol}
\noindent \textbf{Phonon DOS.}
As described in the main manuscript, we evaluate the model performance based on the data splits given in a previous work \cite{chen2021direct}.

\noindent \textbf{Electron DOS.}
On the other hand, for the Electron DOS dataset, we use different dataset split strategies for each scenario.
For the in-distribution setting, we randomly split the dataset into train/valid/test of 80/10/10\%.
On the other hand, for the out-of-distribution setting, we split the dataset regarding the structure of the crystals.
For both scenarios, we generate training sets with simple crystal structures and a valid/test set with more complex crystal structures.
More specifically, in the scenario 1 (different number of atom species, i.e., \# Atom species in Table \ref{tab: main table}), we use binary and ternary crystals as training data and Unary, Quaternary, and Quinary crystals as valid and test data.
In the scenario 2 (different crystal systems, i.e., Crystal System in Table \ref{tab: main table}), we use Cubic, Hexagonal, Tetragonal, Trigonal, and Orthorhombic crystals as training set and Monoclinic and Triclinic as valid and test set.
Please refer to Table \ref{app tab: number of elements} and Table \ref{app tab: crystal system} for detailed statistics for each type of crystal structure.

\noindent \textbf{Fermi Energy.}
We predict the Fermi energy of the crystal structures based on the DOS predicted by the proposed method to evaluate how much physically meaningful DOS is predicted.
To do so, given the ground truth DOS, we first train a four-layered MLP with a non-linearity in each layer to predict the Fermi energy of a crystal structure.
Then, based on the obtained MLP weights, we predict the Fermi energy given the predicted DOS as the input, and calculate the RMSE.
By doing so, we can evaluate how physically meaningful DOS is obtained from each model.

\subsection{Implementation Details}
\label{App: Implementation Details}
In this section, we provide implementation details of \proposed.

{\noindent \textbf{Graph Neural Networks.}}
Our graph neural networks consist of two parts, i.e., encoder and processor.
Encoder learns the initial representation of atoms and bonds,
while the processor learns to pass the messages across the crystal structure.
More formally, given an atom $v_i$ and the bond $e_{ij}$ between atom $v_i$ and $v_j$, node encoder $\phi_{node}$ and edge encoder $\phi_{edge}$ outputs initial representations of atom $v_i$ and bond $e_{ij}$ as follows:
\begin{equation}
    \mathbf{h}^{0}_{i} = \phi_{node}(\mathbf{X}_{i}),~~~ \mathbf{b}^{0}_{ij} = \phi_{edge}(\mathbf{B}_{ij}),
\end{equation}
where $\mathbf{X}$ is the atom feature matrix whose $i$-th row indicates the input feature of atom $v_i$, $\mathbf{B} \in \mathbb{R}^{n \times n \times F_{e}} $ is the bond feature tensor with $F_{e}$ features for each bond.
With the initial representations of atoms and bonds, the processor learns to pass messages across the crystal structure and update atoms and bonds representations as follows:
\begin{equation}
    \mathbf{b}^{l+1}_{ij} = \psi^{l}_{edge}(\mathbf{h}^{l}_{i}, \mathbf{h}^{l}_{j}, \mathbf{b}^{l}_{ij}), ~~~ 
    \mathbf{h}^{l+1}_{i} = \psi^{l}_{node}(\mathbf{h}^{l}_{i}, \sum_{j \in \mathcal{N}(i)}{\mathbf{b}^{l+1}_{ij}}),
\end{equation}
where $\mathcal{N}(i)$ is the neighboring atoms of atom $v_i$, $\psi$ is two layer MLPs with non-linearity, and $l = 0, \ldots, L'$.
Note that $\mathbf{h}^{L'}_{i}$ is equivalent to the $i$-th row of the atom embedding matrix $\mathbf{H}$ in Equation \ref{eq: GNN}.

\noindent \textbf{Model Training.}
In all our experiments, we use the AdamW optimizer for model optimization.
For all the tasks, we train the model for 1,000 epochs with early stopping applied if the best validation loss does not change for 50 consecutive epochs.

\noindent \textbf{Hyperparameter Tuning.}
Detailed hyperparameter specifications are given in Table \ref{app tab: hyperparameter specification}. For the hyperparameters in \proposed, we tune them in certain ranges as follows: number of message passing layers in GNN $L'$ in \{2, 3, 4\}, number of transformer layers for cross attention $L$ in \{2, 3, 4\}, hidden dimension $d$ in \{64, 128, 256\}, learning rate $\eta$ in \{0.0001, 0.0005, 0.001\}, and batch size $B$ in \{1, 4, 8\}.
We use the sum pooling to obtain the crystal $i$'s representation, i.e., $\mathbf{g}_{i}$.
We report the test performance when the performance on the validation set gives the best result.

\begin{table}[h]
\caption{Hyperparameter specifications of~\proposed.}
\centering
\small
\resizebox{0.9 \linewidth}{!}{%
\renewcommand{\arraystretch}{1.1}
\begin{tabular}{cc|ccccc}
\hline
                              &                                                          & \# Message Passing & \# Transformer & Hidden & Learning & Batch \\ 
                              &                                                          & Layers ($L'$) & Layers ($L$) & Dim ($d$) & rate ($\eta$) & Size ($B$)\\ 
                              \hline \hline
\multicolumn{2}{c|}{Phonon DOS}                                                           &                           3&                       2&           256 &             0.001  &     1  \\ \hline
\multicolumn{1}{c|}{\multirow{3}{*}{Electron DOS}} & Random &3  &2  &256  &0.0001  &8  \\
\multicolumn{1}{c|}{} & Scenario 1: \# Atom species  &3  &2  &256  &0.0001  &8 \\ 
\multicolumn{1}{c|}{} &  Scenario 2: Crystal Systems &3  &2  &256  &0.0005  &8 \\ \hline
\end{tabular}}
\label{app tab: hyperparameter specification}
\end{table}

\subsection{Methods Compared}
\label{App: Methods Compared}
In this section, we provide further details on the methods that are compared with \proposed~during experiments.

\smallskip
\noindent \textbf{MLP.}
We first encode the atoms in a crystal with an MLP.
Then, we obtain the representation of crystal $i$, i.e., $\mathbf{g}_{i}$, by sum pooling the representations of its constituent atoms.
With the crystal representation, we predict DOS with an MLP predictor, i.e., $\hat{\mathbf{Y}}^{i} = \phi'(\mathbf{g}_{i})$, where $\phi' : \mathbb{R}^{d} \rightarrow \mathbb{R}^{201}$.

On the other hand, when we incorporate energy embeddings into the MLP, we predict DOS for each energy $j$ with a learnable energy embedding $\mathbf{E}_{j}^{0}$ and obtained crystal representation $\mathbf{g}_{i}$, i.e., $\hat{\mathbf{Y}}^{i}_{j} = \phi(\mathbf{E}_{j}^{0}  + \alpha \cdot \mathbf{g}_{i})$, where $\phi: \mathbb{R}^{d} \rightarrow \mathbb{R}^{1}$ is a parameterized MLP.

\smallskip
\noindent \textbf{Graph Network.}
We first encode the atoms in a crystal with a graph network.
As done for MLP, we obtain the representation of crystal $i$, i.e., $\mathbf{g}_{i}$, by sum pooling the representations of its constituent atoms.
With the crystal representation, we predict the DOS with an MLP predictor, i.e., $\hat{\mathbf{Y}}^{i} = \phi'(\mathbf{g}_{i})$, where $\phi' : \mathbb{R}^{d} \rightarrow \mathbb{R}^{201}$.
Note that the only difference with MLP is that the atom representations are obtained through the message passing scheme.
We also compare the vanilla graph networks with the one that incorporates the energy information by integrating the energy information with initialized energy embeddings as we have done in MLP.

\smallskip
\noindent \textbf{E3NN.}
For E3NN, we use the official code published by the authors\footnote{\url{https://github.com/ninarina12/phononDoS_tutorial}}, which implements equivariant neural networks with E3NN python library\footnote{\url{https://docs.e3nn.org/en/latest/index.html}}.
After obtaining the crystal representation $\mathbf{g}_{i}$, all other procedures have been done in the same manner with other baseline models, i.e., MLP and Graph Network.



\subsection{Additional Experiments}
\subsubsection{Model Performance Analysis}
\label{app: Model Performance Analysis}
In this section, we provide detailed analyses on the model's prediction in the out-of-distribution scenarios.
We have following observations:
\textbf{1)} We observe that \proposed~consistently outperforms in both out-of-distribution scenarios, which  demonstrates the superiority of \proposed.
\textbf{2)} The performance of all the compared models generally degrades as the crystal structure gets more complex. 
That is, models perform worse in Quinary crystals than in Quarternary crystals, and worse in Triclinic crystals than in Monoclinic crystals.
\textbf{3)} On the other hand, it is not the case in Unary crystal. 
This is because only one type of atom repeatedly appears in the crystal structure, which cannot give enough information to the model.
However, \proposed~also makes comparably accurate predictions in the Unary materials by modeling the complex relationship between the atoms and various energy levels.

\begin{table}[H]
\centering
\caption{Model performance in Out-of-Distribution scenarios.}
\resizebox{\textwidth}{!}{%
\begin{tabular}{c|c|cc|cc|cc|cc|cc} \hline
                        \multicolumn{2}{c|}{Data split strategy} & \multicolumn{6}{c|}{Scenario 1: \# Atom species} & \multicolumn{4}{c}{Scenario 2: Crystal System} \\ \hline
\multirow{2}{*} & \multirow{2}{*}{Energy} & \multicolumn{2}{c|}{Unary} & \multicolumn{2}{c|}{Quarternary} & \multicolumn{2}{c|}{Quinary} & \multicolumn{2}{c|}{Monoclinic} & \multicolumn{2}{c}{Triclinic} \\  \cline{3-12}
                        &                                         & RMSE            & MAE           & RMSE            & MAE           & RMSE            & MAE           & RMSE                  & MAE            & RMSE                  & MAE             \\\hline\hline
\multirow{2}{*}{MLP}                     & \multirow{2}{*}{\xmark}                                       & 0.3502               & 0.2992               & 0.2531               & 0.1925               & 0.2850               & 0.2192               & 0.2550                     & 0.1959               & 0.2699                     & 0.2070                 \\
                        &                                        & \scriptsize{(0.0062)}               & \scriptsize{(0.0063)}                & \scriptsize{(0.0027)}                & \scriptsize{(0.0023)}                & \scriptsize{(0.0037)}                & \scriptsize{(0.0032)}                & \scriptsize{(0.0022)}                      & \scriptsize{(0.0022)}                 & \scriptsize{(0.0029)}                      & \scriptsize{(0.0027)}                  \\
\multirow{2}{*}{Graph Network}           & \multirow{2}{*}{\xmark}                                       & 0.2717               & 0.2274               & 0.2124               & 0.1587               & 0.2464               & 0.1847               & 0.1996                     & 0.1485                & 0.2194                     & 0.1649                 \\
                        &                                        & \scriptsize{(0.0039)}               & \scriptsize{(0.0034)}                & \scriptsize{(0.0014)}                & \scriptsize{(0.0019)}                & \scriptsize{(0.0012)}                & \scriptsize{(0.0014)}                & \scriptsize{(0.0010)}                      & \scriptsize{(0.0008)}                 & \scriptsize{(0.0011)}                      & \scriptsize{(0.0013)}                  \\
\multirow{2}{*}{E3NN}                    & \multirow{2}{*}{\xmark}                                       & 0.2013               & 0.1610               & 0.2013              & 0.1443               & 0.2426               & 0.1767               & 0.1805                     & 0.1305                & 0.2021                     &  0.1488               \\
                        &                                        & \scriptsize{(0.0035)}               & \scriptsize{(0.0034)}                & \scriptsize{(0.0009)}                & \scriptsize{(0.0011)}                & \scriptsize{(0.0014)}                & \scriptsize{(0.0012)}                & \scriptsize{(0.0003)}                      & \scriptsize{(0.0004)}                 & \scriptsize{(0.0009)}                      & \scriptsize{(0.0006)}                  \\ \hline
\multirow{2}{*}{MLP}                     & \multirow{2}{*}{\cmark}                                       & 0.2020               & 0.1685               & 0.1997               & 0.1492               & 0.2365               & 0.1782               & 0.1856                     & 0.1406                & 0.2067                     & 0.1578                 \\
                        &                                        & \scriptsize{(0.0007)}               & \scriptsize{(0.0005)}                & \scriptsize{(0.0014)}                & \scriptsize{(0.0009)}                & \scriptsize{(0.0006)}                & \scriptsize{(0.0009)}                & \scriptsize{(0.0007)}                      & \scriptsize{(0.0003)}                 & \scriptsize{(0.0010)}                      & \scriptsize{(0.0008)}                  \\
\multirow{2}{*}{Graph Network}           & \multirow{2}{*}{\cmark}                                       & 0.1913               & 0.1585               & 0.1880               & 0.1379               & 0.2237               & 0.1663               & 0.1709                     & 0.1257                & 0.1911                     & 0.1423                 \\
                        &                                        & \scriptsize{(0.0023)}               & \scriptsize{(0.0021)}                & \scriptsize{(0.0011)}                & \scriptsize{(0.0004)}                & \scriptsize{(0.0016)}                & \scriptsize{(0.0012)}                & \scriptsize{(0.0004)}                      & \scriptsize{(0.0003)}                 & \scriptsize{(0.0013)}                      & \scriptsize{(0.0011)}                  \\
\multirow{2}{*}{E3NN}                    & \multirow{2}{*}{\cmark}                                       & 0.1937               & 0.1562               & 0.1985               & 0.1462               & 0.2383               & 0.1785               & 0.1787                     & 0.1302                & 0.2012                    & 0.1491                 \\
                        &                                        & \scriptsize{(0.0021)}               & \scriptsize{(0.0020)}                & \scriptsize{(0.0005)}                & \scriptsize{(0.0018)}                & \scriptsize{(0.0016)}                & \scriptsize{(0.0023)}                & \scriptsize{(0.0005)}                      & \scriptsize{(0.0005)}                 & \scriptsize{(0.0013)}                      & \scriptsize{(0.0010)}                  \\ \hline
\multirow{2}{*}{\proposed}          & \multirow{2}{*}{\cmark}                                       & \textbf{0.1792}               & \textbf{0.1461}               & \textbf{0.1846}               & \textbf{0.1311}               & \textbf{0.2188}               & \textbf{0.1569}               & \textbf{0.1668}                     & \textbf{0.1188}                & \textbf{0.1880}                     & \textbf{0.1363}                 \\
                       &                                        & \scriptsize{(0.0037)}               & \scriptsize{(0.0034)}                & \scriptsize{(0.0011)}                & \scriptsize{(0.0014)}                & \scriptsize{(0.0008)}                & \scriptsize{(0.0014)}                & \scriptsize{(0.0006)}                      & \scriptsize{(0.0008)}                 & \scriptsize{(0.0020)}                      & \scriptsize{(0.0014)}                  \\ \hline
\end{tabular}%
}
\end{table}


\subsubsection{Model Training and Inference Time}
In this section, to verify the efficiency of \proposed, we compare the training and inference time of the methods during the experiment in Table \ref{tab: complexity}.
\proposed~and E3NN take similar time per training epoch on the Phonon DOS dataset, while E3NN requires much more time per training epoch on the Electron DOS dataset.
This is because the Electron DOS dataset contains a much more diverse and complex crystal structure compared to the Phonon DOS dataset, requiring more time to learn equivariant representations for the structure.
This demonstrates the practicality of \proposed~in real-world applications compared with E3NN.

\begin{table}[h]
    \centering
    \small
    \caption{Training and inference time per epoch for each dataset (sec/epoch).}
    \renewcommand{\arraystretch}{1.2}
    \begin{tabular}{c|c|cc|cc}
    \hline
        & \multirow{2}{*}{Energy} & \multicolumn{2}{c|}{Training} & \multicolumn{2}{c}{Inference} \\ \cline{3-6}
        &  & Phonon DOS & Electron DOS & Phonon DOS & Electron DOS \\ \hline \hline
        MLP & \xmark & 4.12 & 23.79 & 1.36 & 2.66 \\
        Graph Network & \xmark & 16.20 & 59.98 & 1.73 &3.48\\
        E3NN & \xmark & 21.14 & 140.06 &3.61 &9.05  \\ \hline
        MLP & \cmark & 4.73 & 27.74 &1.50 &2.84 \\
        Graph Network & \cmark & 17.37 & 66.72 & 1.95 &3.99\\
        E3NN & \cmark & 22.68 & 149.39 & 3.83 & 9.84\\ \hline
        \proposed & \cmark & 22.09 & 82.00 & 2.11 &4.88 \\ \hline
    \end{tabular}
    \label{tab: complexity}
\end{table}

\end{document}